\documentclass[runningheads]{llncs}
\usepackage[T1]{fontenc}
\usepackage{cite}
\usepackage{graphicx}
\usepackage{amsmath}
\usepackage{multirow}
\usepackage{booktabs}
\usepackage[bb=px]{mathalfa} 

\usepackage{filecontents}
\usepackage{lipsum}

\expandafter\def\expandafter\normalsize\expandafter{%
    \normalsize%
    \setlength\abovedisplayskip{0pt}%
    \setlength\belowdisplayskip{8pt}%
    \setlength\abovedisplayshortskip{-8pt}%
    \setlength\belowdisplayshortskip{2pt}%
}
\begin{document}
\title{Latent Behavior Diffusion for Sequential Reaction Generation in Dyadic Setting}
% \title{Learning Sequential Contexts using Latent Diffusion for Dyadic Reaction Generation\thanks{Supported by BRL \& IITP.}}
% \author{Minh-Duc Nguyen\inst{1}\orcidID{0000-1111-2222-3333} \and
% Hyung-Jeong Yang\inst{2,3}\orcidID{1111-2222-3333-4444} \and
% Third Author\inst{3}\orcidID{2222--3333-4444-5555}\and
% Third Author\inst{3}\orcidID{2222--3333-4444-5555}\and
% Third Author\inst{3}\orcidID{2222--3333-4444-5555}
% }
\author{Minh-Duc Nguyen \and
Hyung-Jeong Yang\inst{*}\and
Soo-Hyung Kim \and
Ji-Eun Shin\and
Seung-Won Kim
}
\authorrunning{Nguyen et al.}
\institute{Chonnam National University, Gwangju, South Korea \\
*Corresponding author e-mail: hjyang@jnu.ac.kr}
% \and
% % Springer Heidelberg, Tiergartenstr. 17, 69121 Heidelberg, Germany
% \email{\{dunm,hjyang,shkim,jieunshin,seungwon.kim\}@jnu.ac.kr}}

% \email{\{dunm,hjyang,shkim,jieunshin,seungwon.kim\}@jnu.ac.kr}}
%
\titlerunning{Latent Behavior Diffusion for Sequential Reaction Generation}
\maketitle              % typeset the header of the contribution
\begin{abstract}
The dyadic reaction generation task involves synthesizing responsive facial reactions that align closely with the behaviors of a conversational partner, enhancing the naturalness and effectiveness of human-like interaction simulations. This paper introduces a novel approach, the Latent Behavior Diffusion Model, comprising a context-aware autoencoder and a diffusion-based conditional generator that addresses the challenge of generating diverse and contextually relevant facial reactions from input speaker behaviors. The autoencoder compresses high-dimensional input features, capturing dynamic patterns in listener reactions while condensing complex input data into a concise latent representation, facilitating more expressive and contextually appropriate reaction synthesis. The diffusion-based conditional generator operates on the latent space generated by the autoencoder to predict realistic facial reactions in a non-autoregressive manner. This approach allows for generating diverse facial reactions that reflect subtle variations in conversational cues and emotional states. 
Experimental results demonstrate the effectiveness of our approach in achieving superior performance in dyadic reaction synthesis tasks compared to existing methods.

\end{abstract}
\section{Introduction}

Dyadic interaction refers to the communication or relationship between two individuals, characterized by direct and reciprocal exchange. This form of interaction is fundamental in social and psychological studies, as it helps to understand interpersonal dynamics, mutual influence, and the development of social bonds. Referring to the Stimulus Organism Response (SOR) model \cite{sor}, each human individual expresses reaction behavior influenced by the context in which they are situated \cite{stimulus}. 
% This context-dependent behavior highlights how environmental, social, and situational factors shape our responses and interactions, emphasizing the complexity and variability of human behavior. 
Specifically, a speaker can significantly affect a listener through various factors such as tone of voice, choice of words, body language, and emotional expressiveness. These elements influence the listener's perceptions, emotions, and responses, thereby shaping the overall communication and interaction dynamics.

% Many previous works have concentrated on speaker modeling, specifically generating talking faces. These efforts have aimed to create lifelike speaker animations by synchronizing lip movements, facial expressions, and speech. Some studies \cite{spk1, spk2, spk3} focus on creating realistic facial animations by directly using input audio to drive the lip movements of the target speaker, leveraging the strong correlation between speech and lip motion. While these efforts have advanced the task of talking face generation, the generation of listener reactions remains largely unexplored. Listener reaction modeling is crucial for realistic digital interactions, as it involves creating responsive facial expressions and head movements that accurately reflect the listener's engagement and emotional responses. Developing such models could significantly enhance applications in human-computer interaction, virtual reality, and animation, providing a more immersive and authentic communication experience. Although the strong correlation between speech and lip motion has been leveraged for speaker modeling, understanding and replicating the nuanced feedback from listeners presents a new and intriguing challenge.

In recent years, there has been an increasing number of studies focusing on the analysis of human-human dyadic interactions \cite{intro0}. These studies aim to understand the intricacies of interpersonal communication by examining verbal and non-verbal cues, emotional exchanges, and the dynamics of social interactions within dyadic contexts. The automated generation of natural facial and bodily reactions, which mimic the behaviors of conversational partners, has been investigated extensively in several studies \cite{intro1, intro2, intro3, intro4, intro5}. These studies primarily focused on replicating specific real facial reactions that correspond to the behavior of the input speaker. However, the potential divergence of non-verbal reaction labels for similar speaker behaviors during the training phase presents challenges for this approach. 

When understanding and replicating the nuanced feedback from listeners presents a new and intriguing challenge, the Responsive Listening Head Generation task was introduced in the computer vision field by Zhou et al. \cite{vico}. Although studies such as \cite{intro1, vico} focused on the nonverbal facial feedback listeners provide to speakers during dyadic conversations, their primary aim was to generate reactions that mirror a ground-truth response and typically employed deterministic models to replicate precise reactions. To capture motion that represents the inherently non-deterministic nature of different perceptually plausible listeners, Learning2Listen \cite{l2l} introduced a framework designed to model interactional communication in dyadic conversations. It processes multimodal inputs from a speaker and autonomously produces multiple potential listener motions in an autoregressive manner, however, the
one-dimensional discrete codebook they used limited the diversity of motion and emotional representation. Later, a study by \cite{dyadic} introduced the novel concept of the Facial Multiple Appropriate Reaction Generation task, pioneering its definition within the literature. 
This study also presented novel objective evaluation metrics tailored to assess the appropriateness of generated reactions. Following the concepts introduced in \cite{dyadic}, this research aims to advance the automatic generation of multiple appropriate non-verbal facial reactions that correspond to specific speaker behaviors.

To tackle this challenge, we propose a novel two-stage, non-autoregressive diffusion architecture for the synthesis of dyadic reactions, also known as Facial Multiple Appropriate Reaction Generation (fMARG). 
% Through extensive experimentation on a real-world dataset, our model has been shown to outperform existing state-of-the-art generative models in generating multiple appropriate facial reactions in response to specific speaker behaviors. 
The primary contributions of our work include:
\begin{itemize}
    \item Leveraging the power of the non-autoregressive Latent Diffusion Model \cite{ldm} as our approach for dyadic reaction generation.
    \item We enhance the latent space representation through a context-aware autoencoder designed to learn spatio-temporal features of the lower facial representation features.
    \item We conduct extensive experiments on the REACT2024 dataset \cite{data4}, demonstrating that our model significantly outperforms recent methods in generating facial reactions.
\end{itemize}

\section{Related Works}
\subsection{Deterministic reaction synthesis}
In recent decades, research on listening reaction modeling has focused on simulating engaged listeners' facial expressions and head movements. Gillies et al. \cite{data-driven} pioneered a data-driven approach to create an animated character capable of dynamically responding to the speaker's voice. Ahuja et al. \cite{head2} focus on generating non-verbal body behaviors. In contrast, Greenwood et al. \cite{head1} explore the synchronized motion of conversational agents in dyadic interactions, with adaptations based on speech. RealTalk \cite{realtalk} utilizes a large language model to retrieve potential videos of the listener's facial expressions. Huang et al. \cite{intro2} trained a conditional Generative Adversarial Network \cite{gan} (GAN) to generate realistic facial reaction sketches of listeners based on the corresponding facial action units (AUs) of the speaker. Song et al. \cite{intro4, intro5} suggest exploring person-specific networks tailored to individual listeners, enabling the reproduction of each listener's unique facial reactions. 

Several studies have explored the generation of diverse non-verbal behaviors, such as hand gestures, posture, and facial reactions, in face-to-face interactions \cite{intro1, ref2}. Zhou et al. \cite{vico} were the first to introduce the Responsive Listening Head Generation task, which involves generating a head video of a listener based on a talking-head video of the speaker and an image of the listener's face. They also developed the ViCo dataset to facilitate the evaluation of methods for this task. Their baseline approach utilized an LSTM-based model \cite{lstm} to process visual and audio data from the speaker to generate facial 3D morphable model (3DMM) \cite{3dmm} coefficients for the listener. Although the former methods could generate listening reaction attributes based on specific speaker behavior inputs, as deterministic models, they lack diversity which is the key to real-world face-to-face scenarios.

\subsection{Multiple Reaction
Generation}
When deterministic approaches grapple with the challenge of the 'one-to-many mapping' problem, where a single speaker behavior can evoke multiple distinct facial reactions, several studies have begun exploring the non-deterministic aspect of this problem, aiming to predict diverse facial reactions from the same input.  Ng et al. \cite{l2l} introduced a novel approach for modeling dyadic communication by predicting multiple realistic facial motion responses from speaker inputs using a motion-audio cross-attention transformer and a motion-encoding VQ-VAE \cite{vqvae} for non-deterministic prediction. This method advances beyond existing work by effectively capturing nonverbal interactions' multimodal and dynamic nature in dyadic conversations. However, expanding the one-dimensional codebook to a composition of several discrete codewords can limit motion and emotional representation diversity. Thus, the Emotional Listener Portrait (ELP) model in \cite{elp}, proposed a discrete motion-codeword-based approach to generate natural and diverse non-verbal responses from listeners based on learned emotion-specific probability distributions and offering controllability.

On the other hand, according to \cite{dyadic}, human facial reactions exhibit variability; identical or similar behaviors from speakers can prompt diverse facial responses, both across different individuals and within the same individual in different contexts. This variability poses challenges when training models to accurately reproduce the listener's facial reactions based solely on each speaker's behavior sequence. Therefore, Song et al. \cite{dyadic, react23, data4} defined the Facial Multiple Appropriate Reaction Generation (fMARG) task and introduced novel objective evaluation metrics to assess the appropriateness of generated reactions. Their framework aims to predict, generate, and evaluate multiple appropriate facial reactions and these models are successful in generating facial responses \cite{1st, 2nd, nguyen2024multiple, nguyen2024vector} by mapping from speaker behavior to a distribution of appropriate reactions. This paper provided an effective approach to the fMARG problem, we designed a Latent Diffusion Model for stochastic listener reaction behavior generation. Our Latent Behavior Diffusion Models accelerate sampling through diffusion in a low-resolution latent space trained by a robust context-aware auto-encoder. This approach achieves state-of-the-art performance in appropriateness, diversity, and synchrony aspects.

% \subsection{Diffusion Generative Models}
% Diffusion Generative Models (DGMs) use an iterative process of adding noise to an initial input to generate high-quality samples, popular for their ability to produce realistic images and sequences.

\section{Proposed Method}
Our method for generating multiple spatio-temporal reactions consists of facial Action Units, valence and arousal intensity, and facial emotion from speaker behavior with the same attributes. Diffusion models are more flexible in how they model data distributions, as they do not rely on adversarial training like GANs, or VAEs, which can suffer from mode collapse. We employ a two-stage process: a context-aware time autoencoder and a Latent Diffusion (LD) generator. The autoencoder updates global statistics iteratively during training to ensure precise reconstruction of future timestamps. Following this, the conditional LD generator utilizes a guidance mechanism to generate latent conditions, effectively incorporating relevant covariates.
\subsection{Problem definition}
Given the $\eta_{th}$ frame ($\eta_{th}\in[t_1, t_2]$ frame) and its preceding frames expressed by the speaker $S_{n}$ at the period $\left[t_{1}, t_{2}\right]$ with corresponding spatio-temporal behaviors $b_{S_{n}}^{t_{1}, t_{2}}$, we develop a generative model $\mathcal{G}$ that predicts each listener appropriate facial reaction frame  $r_{L_{n}}\left(b_{S_{n}}^{t_{1}, t_{2}}\right)_{i}$. This can be formulated as:

% \begin{figure}[t!]
% \centering
% \includegraphics[scale=0.15]{task.png}
% \caption{One-to-many problem in dyadic setting. The same input stimulus from the speaker can elicit different listener reactions and each listener can express different reactions in different contexts.} \label{fig1}
% \end{figure}

\begin{equation}
r_{f}\left(b_{S_{n}}^{t_{1}, t_{2}}\right)_{i}^{\eta}=\mathcal{G}\left(b_{S_{n}}^{t_{1}, t_{2}}\right) \tag{1}
\end{equation}

where $r_{f}\left(b_{S_{n}}^{t_{1}, t_{2}}\right)_{i}^{\eta}$ denotes the $\eta_{\text {th }}$ predicted  facial reaction frame of the $i_{\text {th }}$ generated appropriate facial reaction in response to $b_{S_{n}}^{t_{1}, t_{2}}$; and $b_{S_{n}}^{t_{1}, \eta}$ denotes the speaker behaviour segment at the period $\left[t_{1}, \eta\right]$. Our approach aims to gradually generate all facial reaction frames, resulting in many appropriate spatio-temporal facial reactions (one-to-many) as described
% in Fig.~\ref{fig1}
. Thus, the multiple $M$ appropriate listener facial reaction sequences are presented as:

 \begin{align*}
& R_{f}\left(\text{speaker}^{t_{1}, t_{2}}\right)=\left\{reaction\left(\text{speaker}^{t_{1}, t_{2}}\right)_{1}, \cdots,reaction\left(\text{speaker}^{t_{1}, t_{2}}\right)_{M}\right\}\\
% & P_{f}\left(b_{S_{n}}^{t_{1}, t_{2}}\right)=\mathcal{H}\left(b_{S_{n}}^{t_{1}, t_{2}}\right) \\
& reaction\left(\text{speaker}^{t_{1}, t_{2}}\right)_{1} \neq reaction\left(\text{speaker}^{t_{1}, t_{2}}\right)_{2}^{\eta} \neq \cdots \neq reaction\left(\text{speaker}^{t_{1}, t_{2}}\right)_{M} \tag{2}
\end{align*}

Here, we simply define the set of spatio-temporal listener reaction sequences target $\mathcal{Y}$ and condition speaker behavior $\mathcal{C}$ in the time-space as:

\begin{align*}
\mathcal{Y} &= \left\{ r_{f}\left(b_{S_{n}}^{t_{1}, t_{1}+w}\right)_{M}^{\eta}, \cdots, r_{f}\left(b_{S_{n}}^{t_{2}, t_{2}+w}\right)_{M}^{\eta} \right\}, \quad \mathcal{Y} \in \mathbb{R}^{\mathcal{T} \times d} \tag{3} \\
\mathcal{C} &= \left\{ b_{S_{n}}^{t_{1}-w, t_{1}}, \cdots, b_{S_{n}}^{t_{2}-w, t_{2}} \right\}, \quad \mathcal{C} \in \mathbb{R}^{\mathcal{T} \times d}
\end{align*}

where $\mathcal{T}$ typically represents the number of time steps or temporal observations and
$d$ represents the dimensionality of the features at each time step.
\begin{figure}[t!]
\includegraphics[width=\textwidth]{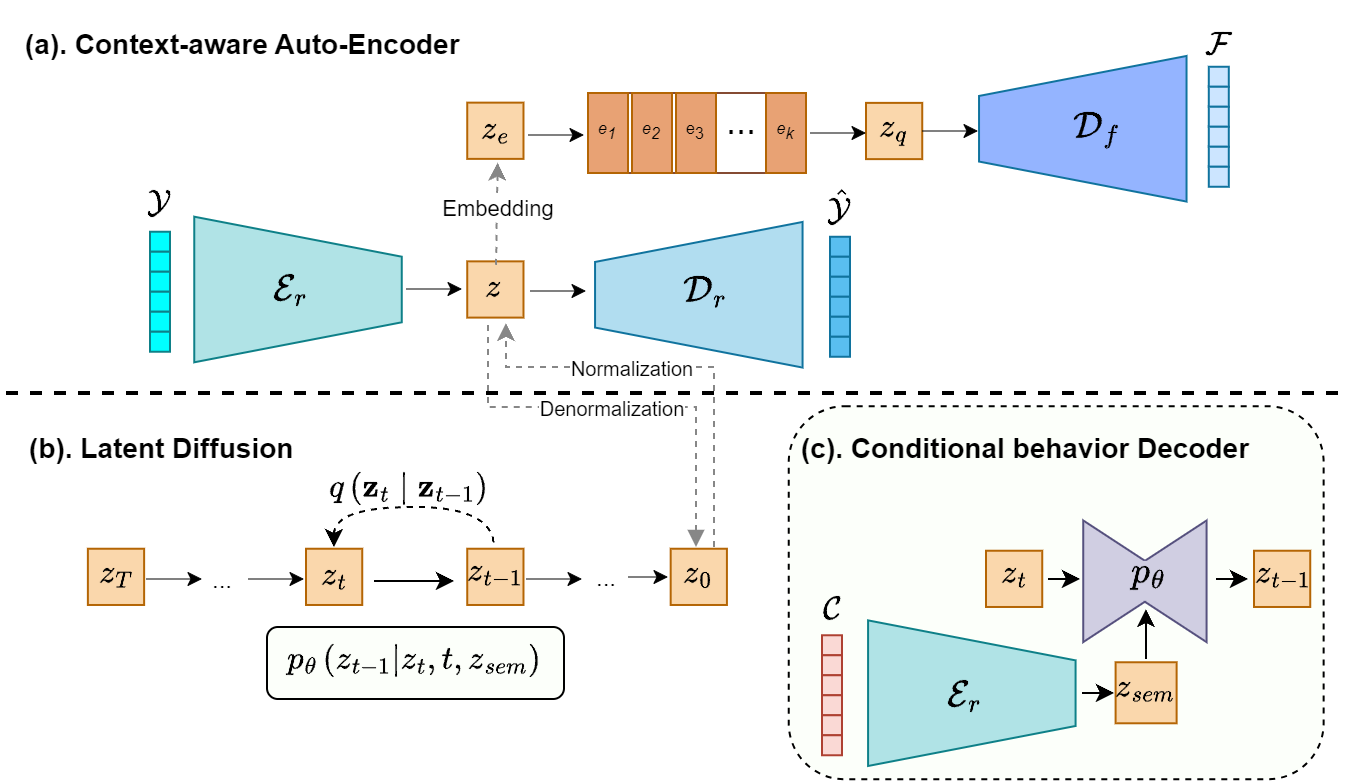}
\caption{Overview of our proposed Latent Diffusion Model (LDM) for generating multiple reactions. During the training phase, the autoencoder (AE) is first trained to encode the time series of listener reactions through a reconstruction task. Concurrently, the LDM is trained to predict future targets based on speaker behaviors $\mathcal{C}$. During the sampling phase, the latent representation of the time series is first generated by the LDM and then passed as input to the decoder 
$\mathcal{D}_{r}(z_0)$ to obtain the future targets.} \label{fig2}
\end{figure}
\subsection{Facial Reaction Compression}
An Auto-Encoder uses backpropagation to generate an output vector similar to its input. It compresses input data into a lower-dimensional space and then reconstructs the original data from this compact representation. In our approach, we implement a context-aware Auto-Encoder (see
Fig.~\ref{fig2}a) with the ability to encode and decode sequences while maintaining awareness of the context or temporal dependencies inherent in the facial reaction sequence data.
% \begin{figure}[t!]
% \centering
% \includegraphics[scale=0.4]{autoencoder.png}
% \caption{The listener reaction Auto-Encoder with deterministic reaction mapping and probabilistic 3D facial features generation.} \label{fig3}
% \end{figure}
\subsubsection{}

The encoder transforms the input $\mathcal{Y}$ into a low-dimensional continuous latent space 
 $V \subset \mathbb{R}^{v}$ 
  (feature code) using a deterministic mapping function: $z=\mathcal{E}_r\left(\mathcal{Y}\right)$. Samples $z \in V$ can be sampled and then reconstructed into the original facial reaction by a decoder $\mathcal{D}_{r}$ as $\hat{\mathcal{Y}} = \mathcal{D}_{r}(z)$. For generating listener 3DMM coordinates, a Vector Quantized technique is performed to produce discrete latent representation $z_q$ by a codebook (see
Fig.~\ref{fig2}a). The 3DMM head motion coefficient set is generated as
$\mathcal{F} = \mathcal{D}_{f}(z_q)$ 

% \begin{align*}
% \left\{\begin{array}{l}
% \boldsymbol{\mu}, \boldsymbol{\sigma}=\mathbf{W}_{\mu} \mathbf{v}, \exp \left(\mathbf{W}_{\sigma} \mathbf{v}\right)  \\
% \mathbf{z}=\boldsymbol{\mu}+\operatorname{diag}(\boldsymbol{\sigma}) \cdot \boldsymbol{\epsilon}, \boldsymbol{\epsilon} \sim \mathcal{N}(\mathbf{0}, \mathbf{I})  
% \end{array}\right. \tag{3}
% \end{align*}

%   \begin{align*}
%   &\hat{b_{f}}\left(b_{S_{n}}^{t_{1}, t_{2}}\right)_{i}^{\delta} = \mathcal{D}(z)\\
% &b_{f}\left(b_{S_{n}}^{t_{1}, t_{2}}\right)_{i}^{\delta} = \{{\hat{r_{f}}\left(b_{S_{n}}^{t_{1}, t_{2}}\right)_{i}^{\delta}, m_{f}\left(b_{S_{n}}^{t_{1}, t_{2}}\right)_{i}^{\delta}} \}\tag{3}
%   \end{align*}

% where $b$ is the predicted listener behavior including reconstructed facial reaction $\hat{r}$ and 3DMM head motion coefficient $m$.  

 % predict the corresponding latent vector $z=\mathcal{E}(\mathbf{Y}) \in V^{2}$

Our Auto-Encoder is trained with a fix-length target listener facial reaction sequence $w$ with Mean Square Error ($L_2$) regularization for $\mathcal{D}_{r}$ and $\mathcal{D}_{f}$ solely. Specifically, we use VQ-VAE loss that is composed of three components: reconstruction loss which optimizes the encoder and decoder; codebook loss to bypass the embedding as the codebook learning by $L_2$ error; and commitment loss to make sure the encoder commits to an embedding for $\mathcal{D}_{f}$.

\begin{align*}
&\mathcal{L}_{react}=\sum_{t=1}^{T}\left\|r_{f}\left(b_{S_{n}}^{t, t+w}\right)_{i}^{\eta}-\hat{r_{f}}\left(b_{S_{n}}^{t, t+w}\right)_{i}^{\eta}\right\|_{2}  \tag{4}
\\  
&\mathcal{L}_{face}=\log p\left(\mathcal{Y} \mid z_{q}(\mathcal{Y})\right)\\
&\quad\qquad+\left\|\operatorname{sg}\left[z_{e}(\mathcal{Y})\right]-e\right\|_{2}^{2}\\
&\quad\qquad+\beta\left\|z_{e}(\mathcal{Y}))-\operatorname{sg}[e]\right\|_{2}^{2}  
\end{align*}

% Where $sg$ represents the stop gradient operator meaning no gradient, $\beta$ denotes the commitment loss hyper-parameter. and $\mathcal{F}_{gt}$ denotes the set of ground truth 3DMM coefficients.
In terms of decoding the listener's spatio-temporal facial reaction, our context-aware Auto-Encoder learns the prior distribution $p(z)$ without a standard Gaussian $\mathcal{N}(0,1)$ or 1-D codebook as it can only produce deterministic outcomes, however, it does not suffer posterior collapse and codebook collapse on a very high-dimension multivariate reaction time series and our posterior Diffusion model can compensate the ability with stochastic inference.

\subsection{Latent Behavior Diffusion}
% \begin{figure}[t!]
% \centering
% \includegraphics[scale=0.4]{diffae.png}
% \caption{A figure caption is always placed below the illustration.
% Please note that short captions are centered, while long ones are
% justified by the macro package automatically.} \label{fig4}
% \end{figure}
We propose a straightforward adaptation of conditional Latent Diffusion Models (LDMs) for reaction generation. 
\begin{figure}[t!]
\centering
\includegraphics[scale=0.35]{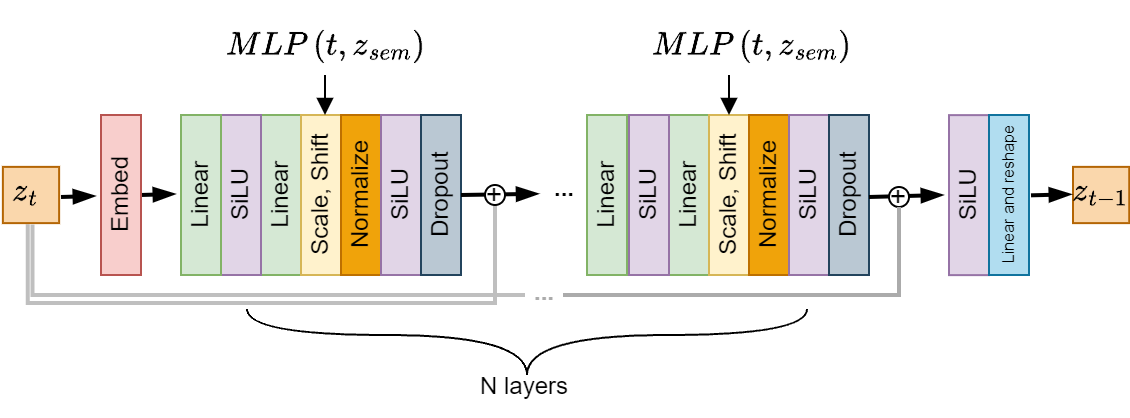}
\caption{Conditional Behavior Decoder architecture in Latent space.} \label{fig3}
\end{figure}
\subsubsection{Conditional behavior Decoder. }In the pursuit of a meaningful latent code, for each denoising step, a sequence of residual MLPs based on Diffusion autoencoders \cite{diffae} is implemented as demonstrated in Fig.~\ref{fig2}c. Each layer of the MLP has a skip connection from
the input, which concatenates the input with the output from the previous layer (see Fig.~\ref{fig3}). Hence, the conditional behavior decoder $p_\theta\left(z_{t-1}|z_{t}, t, z_{sem}\right)$ is conditioned by a semantic behavior encoder $z_{sem} = \mathcal{E}_r(\mathcal{C})$ which reemployed from our First stage Auto-Encoder. In practice, the learned $\mathcal{E}_r$ can deterministic map an
input speaker behaviors $\mathcal{C}$ to a semantically meaningful $z_{sem}$. Here, our conditional behavior decoder takes the high-level semantic subcode $z_{sem}$ and the low-level stochastic subcode $z_T$. In sampling process, our approach 
by reversing the generative process of Pseudo Linear Multi-step (PLMS) \cite{plms} to infer $z_{T}$.
\subsubsection{Diffusion Models. } Diffusion Models \cite{dm} are probabilistic generative models designed to learn the original data distribution 
$P(x)$ by progressively denoising variables sampled from a normal distribution. This process can be viewed as learning the reverse steps of a fixed Markov chain with step length $T \in \mathbb{N}^{+}$. During each step, the diffusion model employs a noise predictor to estimate the noise added in the forward Markov process and then denoise it, effectively refining the data towards its original distribution. The diffusion process is a Markov process, which incrementally applies the forward transition kernel:

% \begin{equation*}
% z_{t+1} \sim \mathcal{N}\left(\sqrt{1-\beta_{t}} z_{t}, \beta_{t} \mathbf{I}\right), t=0,1, \cdots, M-1 \tag{4}
% \end{equation*}

\begin{align*}
& q\left(\mathbf{z}_{1: T} \mid \mathbf{z}_{0}\right)=\prod_{t=T}^{1} q\left(\mathbf{z}_{t} \mid \mathbf{z}_{t-1}\right)  \tag{5}\\
& q\left(\mathbf{z}_{t} \mid \mathbf{z}_{t-1}\right) \sim \mathcal{N}\left(\sqrt{1-\beta_{t}} \mathbf{z}_{t-1}, \beta_{t} \mathbf{I}\right) 
\end{align*}

where $\beta_{t}$ determines the noise strength at each step, referred to as the variance schedule. $T$ represents the total number of steps in the denoising process and $t=0,1, \cdots, T-1$.

The reverse process has a similar form to the diffusion process. During training, the network predicts  $p_{\theta}(z_{t-1}|z_t)$ by
reversing the Markov chain of length M and using a Conditional behavior Decoder $p_\theta(\cdot)$ presented as:

\begin{align*}
& p\left(\mathbf{z}_{0: T}\right)=  p\left(\mathbf{z}_{T}\right) \prod_{t=t}^{T} p_{\theta}\left(z_{t-1}|z_{t}, t, z_{sem}\right)
\tag{6}\\
&p\left(\mathbf{z}_{T}\right)=\mathcal{N}(0, \mathbf{I}) \\
&  q\left(\mathbf{z}_{t-1} \mid \mathbf{z}_{t}, \mathcal{Y}\right)= \mathcal{N}\left(\mu_{t}\left(\mathbf{z}_{t}, \mathcal{Y}\right), \sigma_{t}^{2} \mathbf{I}\right)
\end{align*}

where $\mu_{t}\left(\mathbf{z}_{t}, \mathcal{Y}\right)$ has a closed-form  solution and $\sigma_{t}$ is a hyperparameter.

\subsubsection{Objective Functions. }The LDM’s loss is the average of the Mean Absolute Error (MAE) in the latent space and the Mean Square Error (MSE) in the reconstructed space. The complete objective training loss function for our LDM is expressed as follows:

\begin{align*}
& \mathcal{L}_{latent}=\sum_{t=1}^{T} \underset{q\left(z_{t} \mid z_{0}\right)}{\mathbb{E}}\left\|p_\theta\left(z_{t-1}|z_{t}, t, z_{sem}\right)-\mathcal{E}_r(\mathcal{Y})\right\|_{1} \ \tag{7}
\\
& \mathcal{L}_{rec}=\sum_{t=1}^{T} \underset{q\left(z_{t} \mid z_{0}\right)}{\mathbb{E}}\left\|\mathcal{D}_r(p_\theta\left(z_{t-1}|z_{t}, t, z_{sem}\right))-\mathcal{Y}\right\|_{2} 
\\
& \mathcal{L}_{ldm} =  \mathcal{L}_{latent} + \mathcal{L}_{rec}
\end{align*}

\subsubsection{Latent Behavior Sampler. }
Pseudo Linear Multi-Step \cite{plms} (PLMS) is an improvement over DDIM
. According to \cite{plms}, a 50-step process can achieve higher quality than a 1000-step process in DDIM. We proposed a Latent Behavior Sampler that can be broken down into a series of formulas capturing the core logic of sampling $z_{t+\delta}$ from the model using the PLMS method. Following the forward Euler Method, for a certain differential equation satisfying $\frac{d x}{d t}=f(z, t)$. 
We represent PLMS in formulaic terms:

\begin{align*}
&z_{t+\delta}=z_{t}+\delta\left(f_{t}\right) &  \text{if } k=1
\\ \tag{8}
&z_{t+\delta}=z_{t}+\frac{\delta}{2}\left(3 f_{t}- f_{t-\delta}\right) &  \text{if } k=2
\\
&z_{t+\delta}=z_{t}+\frac{\delta}{12}\left(23 f_{t}-16 f_{t-\delta}+5 f_{t-2 \delta}\right) &  \text{if } k=3
\\
&z_{t+\delta}=z_{t}+\frac{\delta}{24}\left(55 f_{t}-59 f_{t-\delta}+37 f_{t-2 \delta}-9 f_{t-3 \delta}\right) &  \text{if } k=4
\\
& 
\end{align*}

where $f_{t}=f\left(z_{t}, t\right)$, $k$ is the convergence order and $\delta$ is the step size. Our inference process leverages the reverse PLMS process when it reaches $T=0$, obtaining the future target by $\mathcal{D}_r(z_0)$.

% \begin{cases}k_{1}=f\left(x_{t}, t\right) & k_{2}=f\left(x_{t}+\frac{\delta}{2} k_{1}, t+\frac{\delta}{2}\right)  \tag{6}\\ k_{3}=f\left(x_{t}+\frac{\delta}{2} k_{2}, t+\frac{\delta}{2}\right), & k_{4}=f\left(x_{t}+\delta k_{3}, t+\delta\right) \\ x_{t+\delta}=x_{t}+\frac{\delta}{6}\left(k_{1}+2 k_{2}+2 k_{3}+k_{4}\right)
% \end{cases}

\section{Experiments}
\subsection{ Evaluation setup}
\noindent\textbf{Dataset.}
The REACT2024 dataset \cite{data4, dyadic} is a comprehensive resource for analyzing dyadic video interactions with detailed facial attribute annotations. This dataset is constructed from two prominent video conference corpora: NoXi \cite{noxi} and RECOLA \cite{recola}. Each audio-video clip from the NoXi and RECOLA datasets has been segmented into 30-second clips, resulting in conversational 5,919 clips. 
% Most of these clips, 5,870 (49 hours), are sourced from the NoXi dataset, while 54 clips (0.4 hours) come from the RECOLA dataset. 
Specifically, the dataset includes extensive facial attribute annotations for each frame, derived using state-of-the-art models \cite{data1, data2, data3}:
Action Units (AUs): 15 AUs are annotated, including AU1, AU2, AU4,... 
% AU6, AU7, AU9, AU10, AU12, AU14, AU15, AU17, AU23, AU24, AU25, and AU26;
Facial Affects: Two continuous affective states, valence and arousal intensities, are provided; Facial Expression Probabilities: Eight facial expression probabilities are included, covering Neutral, Happy, Sad,... and further the extracted 3DMM parameters.
% Surprise, Fear, Disgust, Anger, and Contempt. 
The carefully segmented and cleaned clips, rich annotations, and special appropriateness label strategy provide a robust foundation for training and evaluating advanced machine-learning models in dyadic emotion recognition, facial expression analysis, interaction dynamics in video conferencing scenarios, and indeed capability for reaction generation.

\noindent\textbf{Comparison Methods.} We compare our method with two baseline approaches provided by \cite{react23, data4}: TransVAE and Belfusion. The Trans-VAE baseline shares a similar architecture to the TEACH model proposed in \cite{transvae}, a Multimodal Transformer-based VAE that takes video facial and audio embedding from speaker to predict listener facial reaction features and 3DMM parameters. Belfusion is based on the work in \cite{belfusion}, employing DDIM model with standard Gaussian Distribution as the prior. Furthermore, some generative methods were remarkable at REACT2024 Challenge \cite{data4} included in our comparison. In particular, Dam et al. \cite{1st} designed an architecture encouraged by \cite{l2l}, but leveraging Finite Scalar Quantization \cite{finite} to replace Vector Quantization. Besides, Liu et al. \cite{2nd} introduced discrete latent variables to tackle this one-to-many mapping problem, to model the diversity of contextual factors, and to generate diverse reactions.

\noindent\textbf{Implementation Details.}
We implement our model using PyTorch \cite{torch} and perform the training
on a single Nvidia RTX 3080Ti GPU. In the first stage of our autoencoder structure, we implemented based on Transformer-based \cite{trans} architectures, all $\mathcal{E}_r$, $\mathcal{D}_f$, and $\mathcal{D}_r$ utilized two layers of Transformer encoders, each with 4 attention heads. The latent Conditional Behavior Decoder includes MLP + Skip with 10 layers and 1024 hidden nodes, detailed in Fig.~\ref{fig3}.
The AE models are solely trained with 1000 epochs for $\mathcal{D}_r$ and 200 epochs for $\mathcal{D}_f$ with a batch size of 32. The window size is set to 50, the learning rate is 1e-3, and the weight decay is 5e-4 with the AdamW optimizer. We adjusted the same optimizer parameters for the second stage and trained the LDM for 200 epochs. The denoising
chain has 50 steps. Sampling was conducted with our fourth-order PLMS latent behavior sampler.

\subsection{Evaluation metric}
Following the standard protocols proposed in \cite{dyadic}, we evaluate our method based on three key aspects of the generated facial reaction attributes as below:

\noindent\textbf{Appropriateness:} Facial reaction distance (FRDist) calculates the Dynamic Time Warping (DTW) distance between a generated facial reaction and its closest corresponding real facial reaction; Facial Reaction Correlation (FRCorr) computes the correlation between each generated facial reaction and its most similar corresponding real facial reaction.
      \\
\noindent\textbf{Diversity:}Facial Reaction Variance (FRVar) computing the variation across all frames; Diverseness among generated facial reactions (FRDiv); Diversity among facial reactions generated from different speaker behaviors (FRDvs).\\
% Facial Reaction Variance (FRVar) computing the variation across all frames; Diverseness among generated facial reactions (FRDiv) is the sum of the Mean Squared Error (MSE) among every pair of generated facial reactions in response to each input speaker behavior; Diversity among facial reactions generated from different speaker behaviors (FRDvs) evaluates the diversity of responsive reactions to different speaker behaviors.
\noindent \textbf{Synchrony} (FRSyn) is computed by first calculating the Time-Lagged Cross-Correlation (TLCC) scores between the input speaker behavior and each of its generated facial reactions.
\begin{table*}[t!]
    \centering
    \caption{Comparision against various methods on Multiple Facial Reaction Generation on REACT2024 dataset. }
    \renewcommand{\arraystretch}{1.5}
    \begin{tabular}{lcccccc}
    \hline
\multirow{2}{5em}{Method}&  \multicolumn{2}{c}{Appropriateness} & \multicolumn{3}{c}{Diversity}& Synchrony  \\ \cmidrule(l){2-3} \cmidrule(l){4-6}  \cmidrule(l){7-7}
         &FRCorr$(\uparrow)$    &FRDist$(\downarrow)$   &FRDiv$(\uparrow)$   &FRVar$(\uparrow)$  &FRDvs$(\uparrow)$  &FRSyn$(\downarrow)$ \\   \hline
         Trans-VAE&  0.07&  90.31&  0.0064 &  0.0012&  0.0009& 44.65 \\ 
         BeLFusion& 0.12 & 94.09 & 0.0379  & 0.0248 & 0.0397 & 49.00\\ 
         Dam et al. \cite{1st}& 0.31 & \textbf{84.94} & 0.1167  & 0.0349 & 0.1165 & 47.43\\ 
        Liu et al. \cite{2nd} & 0.22 & 88.32 & 0.1030 & 0.0387 & 0.1065 & 44.41\\ \hline
        % Ours\_T& \textbf{0.39} & 94.82 & \textbf{0.12578} & \textbf{0.06518} & \textbf{0.14933} & 45.66\\ \hline
         Ours&\textbf{0.37} & 89.40 & \textbf{0.1211}  & \textbf{0.0653} & \textbf{0.1505}& \textbf{43.48}\\ \hline
   
    \end{tabular}

    \label{tab:1}
    % {\footnotesize Note: k is the number of denoise steps.}
    % \parbox{4.6in}{
% \footnotesize Note: k is the number of denoise steps.}
\end{table*}
\begin{table*}[t!]
\centering
\caption{Comparisons on image-level in terms of quality and identity
preservation.}\label{tab:2}
\begin{tabular}{ll}
\hline
Method &  FID $(\downarrow)$ \\
\hline
GT &  { 53.96} \\
Trans-VAE &  { 69.19} \\
Belfusion &  { 54.00} \\
Ours & { \textbf{50.95}} \\
\hline
\end{tabular}
\end{table*}
\subsection{Results}

\noindent\textbf{Quantitative Results.} Table \ref{tab:1} illustrates the quantitative evaluation results of our proposed LDM compared with other methods. The results demonstrate that our approach generates facial reactions with greater diversity and synchrony than those produced by the competitors. Trans-VAE and Belfusion got the worst performance among our comparison, these models relied on standard Gaussian distribution can easily lead to posterior collapse during training with Trans-VAE and prior VAE of Belfusion. Dam et al. \cite{1st} and Liu et al. \cite{2nd} tackled this problem by applying discrete latent space.
Nevertheless, our Latent Diffusion model with a Transformer AE prior can surpass the overall evaluation criteria, delivering superior performance in generating diverse and synchronized facial reactions. 
% In addition, the Latent Diffusion model with Transformer-based AE achieved the best scores in diversity and FRCorr metrics.
Moreover, the better diversity and synchrony metrics state that our Decoder can better generate responsive reactions and avoid jitters in the facial frames among generated fix-length segments.
Although there was a trade-off regarding FRDist, its performance remains competitive and within a fair margin compared to other approaches. 
% This can be explained by the fact that our LSTM Decoder can better avoid jitters in the facial frames among generated fix-length segments compared to a Transformer Decoder.

To comprehensively evaluate the video-level performance, we use the Fréchet Inception Distance (FID) \cite{fid} as a commonly used metric for assessing the realism of generated human faces. We evaluate the realism of the facial reactions that PIRender \cite{pirender} generated from 3DMM parameters by $\mathcal{D}_{f}$,  results are shown in Table~\ref{tab:2}.

\begin{figure}[t!]
\includegraphics[width=\textwidth]{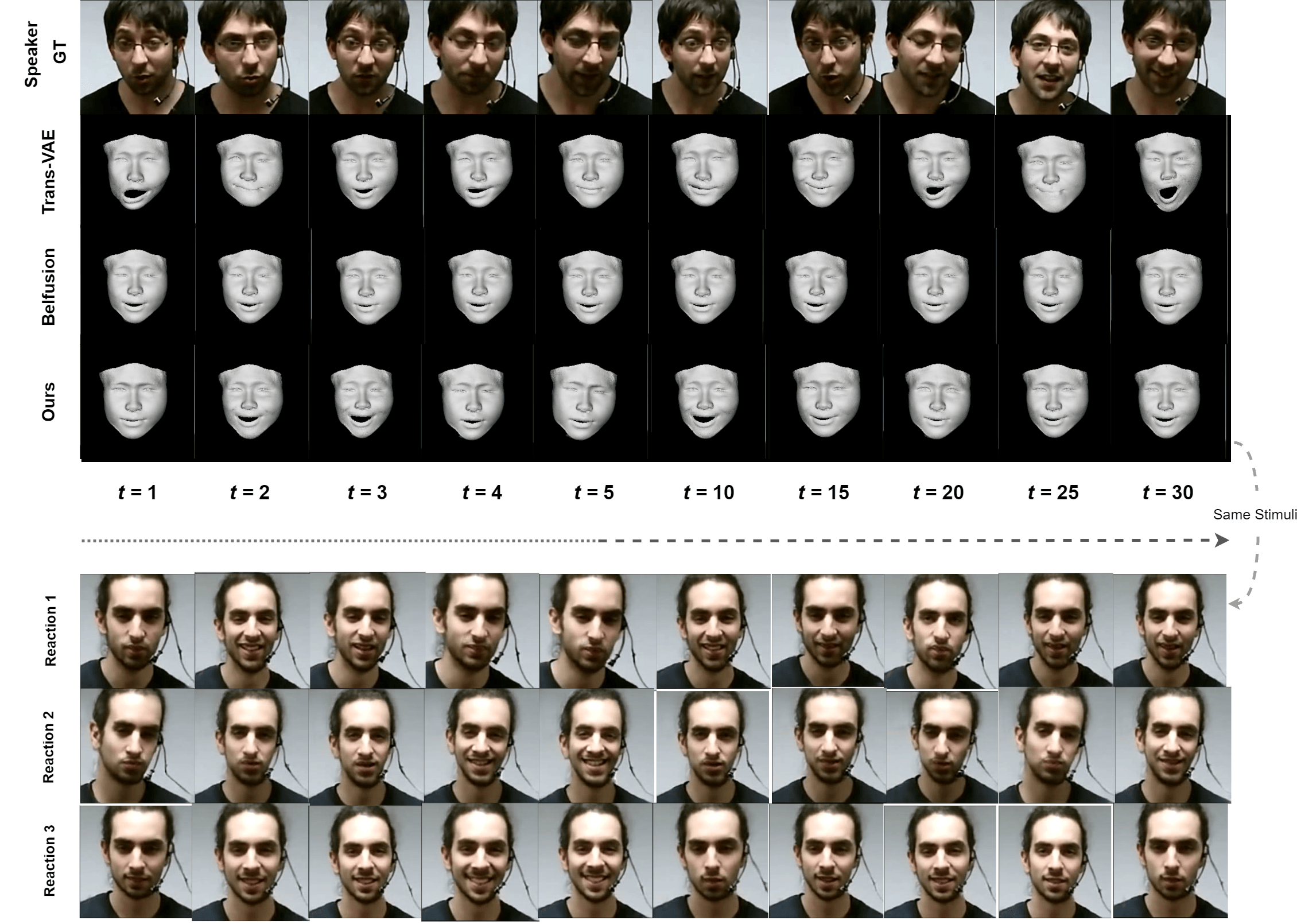}
\caption{The qualitative results of
the generated facial frames of multiple listeners. Our comparison between other baselines by 3D rendering translation. The bottom of the figure shows our model-generated variants of reaction that are expressed from Speaker ground truth. The time $t$ in second.} \label{fig5}
\end{figure}

\begin{figure}[t!]
\centering
\includegraphics[scale=0.25]{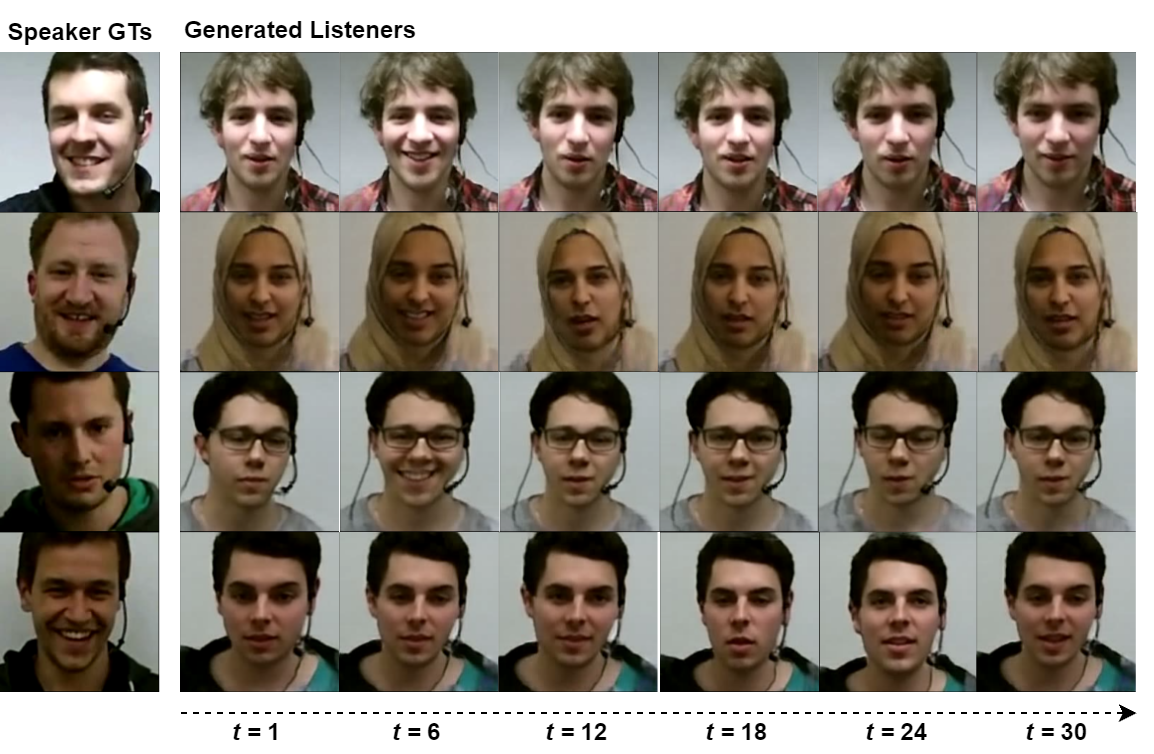}
\caption{Generated listener facial reactions from different speakers.} \label{fig6}
\end{figure}

\noindent\textbf{Qualitative Results.} In this section, we present the qualitative results of the generated facial frames of multiple listeners. These results are illustrated in Fig.~\ref{fig5} and Fig.~\ref{fig6}. Trans-VAE failed to preserve identity, generating unclear expressions and arbitrary motions. Belfusion's results show poor variation and less dynamic motions due to the weakness of its Gaussian latent space. In contrast, our model achieves the best natural and coherent results compared to these baseline methods.

% \begin{tabular}{|c|c|c|c|c|c|c|}
% \hline & \multicolumn{2}{|c|}{ Appropriateness } & \multicolumn{3}{|l|}{ Diversity }  & \multirow{2}{*}{\begin{tabular}{l} 
% Synchrony \\
% FRSyn ($\cdot$) \\
% \end{tabular}} \\
% \hline & FRCorr $(\uparrow)$ & FRDist $(\downarrow)$ & FRDiv $(\uparrow)$ & FRVar $(\uparrow)$ & $\overline{\text { FRDvs }(\uparrow)}$ & & \\
% \hline Ground truth & 8.73 & 0.00 & 0.0000 & 0.0724 & 0.2483  & 47.69 \\
% \hline Random & 0.05 & 237.23 & 0.1667 & 0.0833 & 0.1667 &  44.10 \\
% \hline Mime & 0.38 & 92.94 & 0.0000 & 0.0724 & 0.2483 &  38.54 \\
% \hline MeanSeq & 0.01 & 97.13 & 0.0000 & 0.0000 & 0.0000 &  45.28 \\
% \hline MeanFr & 0.00 & 97.86 & 0.0000 & 0.0000 & 0.0000 &  49.00 \\
% \hline Trans-VAE & 0.07 & 90.31 & 0.0064 & 0.0012 & 0.0009 &  44.65 \\
% \hline BeLFusion & 0.12 & 94.09 & 0.0379 & 0.0248 & 0.0397 &  49.00 \\
% \hline$\overline{\text { Ours }}$ & 0.31 & 84.93 & 0.1164 & 0.0348 & 0.1166 &  47.42 \\
% \hline
% \end{tabular}


\begin{table*}[t!]
    \centering
    \caption{Effect of ablating key settings of our method.}
    \renewcommand{\arraystretch}{1.5}
    \begin{tabular}{lcccccc}
    \hline
\multirow{2}{5em}{Method}&  \multicolumn{2}{c}{Appropriateness} & \multicolumn{3}{c}{Diversity}& Synchrony  \\ \cmidrule(l){2-3} \cmidrule(l){4-6}  \cmidrule(l){7-7}
         &FRCorr($\uparrow$)    &FRDist($\downarrow$)   &FRDiv($\uparrow$)   &FRVar($\uparrow$)  &FRDvs 
 ($\uparrow$)&FRSyn($\downarrow$)  \\ \hline
         PLMS (k=2, T=10)& 0.36&  89.27&  0.1157 &  0.0627& 0.1434&44.16 \\ 
         PLMS (k=2, T=25)& 0.37 & 91.99 & 0.1190 & 0.0699 & 0.1605 & 45.47 \\ 
         PLMS (k=1, T=50)& 0.372 & 89.49 & 0.1211 & 0.0653 & 0.1505 & 43.88\\ 
         PLMS (k=2, T=50)& 0.372 & 89.48 & 0.1206  & 0.0651 & 0.1501& 43.68 \\
         PLMS (k=3, T=50)& 0.372 & 89.54 & 0.1208 & 0.0651 & 0.1501 &  43.69\\
          PLMS (k=4, T=50)& \textbf{0.374} & 89.40 & \textbf{0.1211}  & \textbf{0.0653} & \textbf{0.1505}& \textbf{43.48} \\ 
       PLMS (k=2, T=100)& 0.33 & 88.22 & 0.1064 & 0.0619 & 0.1405&45.03 \\ \hline
        DDIM (T=50)& 0.35 & 88.27 & 0.1047 & 0.0579 & 0.1343&44.00 \\ 
         DPM (T=50)& 0.32 &  \textbf{86.53} & 0.1014 & 0.0546& 0.1223&43.82\\ \hline
   
    \end{tabular}
    
    \label{tab:3}
    % {\footnotesize Note: k is the number of denoise steps.}
    \parbox{4.6in}{
\footnotesize Note: T is the number of denoise steps and k is convergence order of PLMS.}
\end{table*}
\subsection{Ablation Study}
We quantitatively analyze the effect of each of the settings in our method for the final LD model with Trans-AE prior. We perform ablation studies with different denoise chain step numbers $T \in{(10, 25, 50, 100)}$ and four convergence orders of PLMS method. 
% We further add DDIM and the origin DPM samplers as other sampling methods in our comparison.

Table \ref{tab:3} demonstrates that increasing the number of denoising steps slightly improves the alignment of predicted reactions with the ground truth in terms of Dynamic Time Warping (DTW). However, this improvement negatively impacts the Correlation aspect. Through our experiments, we identified an optimal balance between Appropriateness and Diversity at 50 denoising steps. The impact of varying PLMS convergence orders is minimal, with our best results achieved using the fourth-order implementation. Furthermore, we applied DDIM and the origin DPM samplers with the same 50 denoising steps, and our PLMS achieved higher Correlation and Diversity in our comparison.

\section{Conclusion}
In this paper, our study introduces Latent Behavior Diffusion as a robust framework for generating diverse and contextually appropriate facial reactions in dyadic interactions. By combining a context-aware autoencoder with a diffusion-based conditional generator, our approach effectively compresses high-dimensional input features into a concise latent representation. Our approach attains superior performance in objective benchmarks of generated facial reactions compared to existing methods. We further aim to improve the appropriateness of DTW by refining denoising process for better long-term reaction sequence modeling.

\subsubsection{Acknowledgements} This work was supported by the National Research Foundation of Korea (NRF) grant funded by the Korea government (MSIT) (RS-2023-00219107). This work was also supported by the Institute of Information \& communications Technology Planning \& Evaluation (IITP) under the Artificial Intelligence Convergence Innovation Human Resources Development (IITP- 2023-RS-2023-00256629) grant funded by the Korean government (MSIT).

\nocite{*}
\bibliographystyle{unsrt}
\bibliography{annot}

\end{document}